\documentclass[aps,prl,twocolumn,superscriptaddress]{revtex4-2}

\usepackage{amsmath}
\usepackage{amssymb}
\usepackage{graphicx}
\usepackage{hyperref}
\usepackage[table]{xcolor}

\newcommand{\bbeta}{\boldsymbol{\beta}}
\newcommand{\bb}{\boldsymbol{b}}

\usepackage{amsthm}
\newtheorem{theorem}{Theorem}
\newtheorem{lemma}{Lemma}
\usepackage{placeins}

\begin{document}
\let\origaddcontentsline\addcontentsline
\renewcommand{\addcontentsline}[3]{}

\title{Phase Transitions in Learning Driven by Extremal Data}

\author{Ethan Levien}
\email{ethan.a.levien@dartmouth.edu}
\affiliation{Department of Mathematics,
Dartmouth College, Hanover, NH 03755, USA}

\author{Trevor GrandPre}
\email{trevorg@wustl.edu}
\affiliation{Department of Physics,
Washington University in St. Louis, St. Louis, MO 63130, USA}
\affiliation{National Institute for Theory and Mathematics in Biology,
Northwestern University and The University of Chicago, Chicago, IL, USA}

\date{\today}

\begin{abstract}
We study an extremal regression problem, an example of multiple instance learning (MIL), where the predictors are given as multisets, called bags. The bag elements (called instances) possess hidden labels which are pooled to produce an observed bag-level label. Max pooling is common in MIL applications, but gives rise to vanishing gradients. To handle this, modern MIL pipelines approximate the max with a softmax, treating the sharpness as a learnable parameter, denoted $\lambda$. When the max pooling model is the true data generating process, learning will be successful when $\lambda$ is increased during gradient descent, thereby driving the dynamics into regions of parameter space with vanishing gradients. To understand the conditions under which gradient descent breaks down, we study a minimal model and derive a low dimensional projection of the gradient flow in the joint limit of large bags $n$ and high dimension $d$ with $\alpha=\ln n/d$ fixed. The flow has two distinct phase transitions: an intrinsic freezing transition related to the random energy model of disordered systems, and a retrieval transition occurring in modern Hopfield networks.  Numerical simulation confirms both the phase diagram and the convergence of the finite-size training gradient to its predicted asymptotic form. \end{abstract}

\maketitle

\textit{Introduction.---}

Many supervised learning problems in the natural sciences can be formulated as having set valued inputs (sets of images, pieces of text, immune repertoire sequences) where the labels are determined by a small fraction, or even a single 
element in the set. For example, the strongest binding antibody sequence or image containing an object of interest. These are examples of Multiple Instance Learning (MIL)
\cite{dietterich1997solving,Carbonneau2018survey,jain1994compass,sliwoski2014computational,zankov2024chemical}.
In the MIL literature, the sets are called bags and the elements of the bags are called instances. A label is assigned at the bag level by pooling over latent instance-level labels or scores \cite{dietterich1997solving,maron1998framework,maron1998learning}.
Learning in MIL problems is challenging because the instance level labels are unobserved and the mapping from instance level to bag level labels is typically highly nonlinear. 

\begin{figure}[t]
  \centering
  \includegraphics[width=\columnwidth]{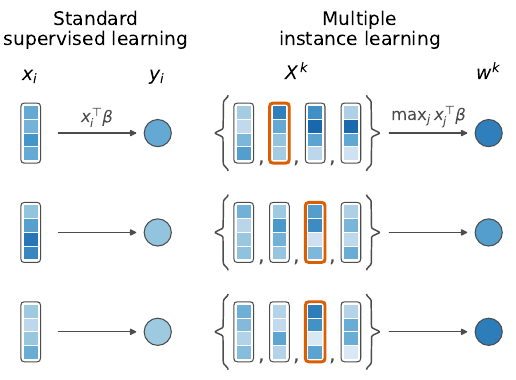}
  \caption{Standard supervised learning maps each feature vector $x_i$ to
    a label $y_i = x_i^\top\beta$. In MIL, bags $X^k$
    of instances are mapped to a single label $w^k$ produced by an unobserved
    selection rule (e.g. the max). The goal is to learn
    $\beta$ from $(w^k, X^k)$ pairs alone without ever observing which
    instance determined the label.}
  \label{fig:intro}
\end{figure}

We consider real valued bags, represented as matrixes $X^k\in\mathbb{R}^{n\times d}$ whose rows 
$\boldsymbol x_i\in\mathbb{R}^d$ are the instances, and take the \emph{true} labels to be generated via 
\begin{equation}
  w^k = \max(\boldsymbol y^k), \qquad \boldsymbol y^k = X\bbeta_{\rm true}.
  \label{eq:mlf}
\end{equation}
Figure~\ref{fig:intro} depicts an example. 
We assume that some bags are ``positive,'' meaning they contain the target direction
$\bbeta_{\rm true}$ in feature space, while others are sampled from a background distribution.  In applications, such instances might correspond to a
nondiagnostic tissue patch or a molecular conformation that does not
mediate binding.  To learn $\bbeta_{\rm true}$, the max is approximated with a softmax whose sharpness, $\lambda$ is itself learnable. When does gradient descent actually recover $\bbeta_{\rm true}$?

Using the REM correspondence of
\cite{lucibello2024exponential,mezard2009information}, we obtain a
low-dimensional description of the gradient descent dynamics as
$n,d\to\infty$. This dynamics exhibits two phase transitions: One
is a freezing transition inherited from the REM and the other is
a retrieval transition arising from competition between the growth of
the softmax weight on the planted instance in positive bags and the sum
over the other weights. 

As a result of these phase transitions the gradient dynamics become insensitive to the angle between the learned direction and $\bbeta_{\rm true}$ until the dynamics
enters the retrieval phase. Separately, the gradient of $\lambda$ vanishes upon reaching
the low-temperature branch of the REM transition. We show numerically both that
this phase diagram of the limit system accurately predicts key attributes of the finite-size training trajectories and
that the finite-$(n,d,\ell)$ training gradient itself converges to the
asymptotic gradient-flow field we derive (Supplemental Material,
Sec.~\ref{si:sec:numerics}).

\textit{Related work.---}%
MIL was formalized by Dietterich \emph{et al.} \cite{dietterich1997solving} and Maron and Lozano-P\'erez presented a
general framework built on the notion of a hidden \emph{concept point}
\cite{maron1998framework,maron1998learning}. Early algorithmic approaches
include diverse-density/EM-DD \cite{zhang2002emdd}, MI-SVM
\cite{Andrews2002MISVM}, and multi-instance kernels
\cite{Gartner2002MIKernels}. Theoretical results for learnability of classification MIL problems have been studied in 
\cite{diochnos2012multiple,doran2016multiple,szabo2015two}. We study
multiple-instance regression specifically, which was studied 
beginning with Ray and Page \cite{RayPage2001MIR} and Wagstaff and Lane
\cite{WagstaffLane2007StructuredMIR} and subsequently extended in
\cite{dooly2002multiple,dooly2006real,Chauhan2024GenLearnMIR,Levien2026EMQueryValue}.
Softmax-smoothed pooling with a learnable temperature was introduced in 
\cite{ilse2018attention} and applied to immune repertoire classification 
in \cite{widrich2020deeprc}. Softmax pooling is also at the core of self-attention layers in transformers
\cite{Vaswani2017} and of modern Hopfield networks
\cite{ramsauer2020hopfield,hopfield1982neural}. The storage capacity for
continuous pattern retrieval in Hopfield networks was analyzed via a random-energy-model (REM)
correspondence by Lucibello and M\'ezard \cite{lucibello2024exponential},
building on the classical REM
\cite{derrida1980random,derrida1981random,mezard2009information} and its
sum-of-exponentials limit theorems \cite{ben2005limit}. Supplemental
Material Sec.~\ref{si:sec:hopfield} makes the correspondence between our retrieval
transition and pattern retrieval in these models precise. See also
\cite{vuckovic2020mathematical} for a general mathematical treatment of
attention.

\textit{Data generating process.---}%
We consider a dataset $\{w^k,X^k\}_{k=1}^\ell$ with
$X^k\in\mathbb R^{n\times d}$ and $w^k$ given by Eq. \eqref{eq:mlf}. 
The
background population is modeled iid Gaussian, $\boldsymbol x_i\sim{\rm Normal}(0,I_d)$. For a
fraction $q$ of bags, one instance is of 
$\bbeta_{\rm true}$ is then planted in the bags.  
It will be useful to separate the direction of the learned regression vector from its norm; thus, we write $\bbeta_{\rm true}=\rho_{\rm true}\bb_{\rm true}$ and the learned regression vector as 
$\bbeta=\rho\bb$ with $\|\bb\|^2=\|\bb_{\rm true}\|^2=d$.  

Motivated by real MIL data sets where there can be hundreds of thousands of instances per bag, we will take the large $n$ limit. The dimension $d$ is often controlled by the model architecture; specifically, how the raw data such as image patches or sequences are embedded as real vectors. For iid Gaussian instances, this embedding dimension must scale with the bag size for the $\bbeta_{\rm true}$ to be learnable, and the scaling is dictated by the order of the extremes, since this competes with the planted instances.  A basic result in the theory of extremal statistics states that for $n$ iid standard Gaussian $x_i$, $(\ln n)^{-1/2}\max_i x_i  \to \sqrt{2}$ almost surely \cite[Lemma 9.1.1]{bovier2006statistical}. In our notation, $\langle \boldsymbol x_i,\bb \rangle \sim{\rm Normal}(0,d)$, motivating us to define $\alpha=\ln n/d$, so that 
\begin{equation}\label{eq:exlim-as}
\langle \boldsymbol  x_i, \bb \rangle/d \to \sqrt{2\alpha}.
\end{equation}
The positive bags are identifiable (i.e.\ the
planted instance realizes the true maximum) only when
$\alpha=\ln n/d<\rho_{\rm true}^2/2$. We work in this identifiable regime
throughout thereby isolating questions about what happens during gradient descent.  

\textit{Training.---}%
The is the central
obstruction to fitting $\bbeta$ by gradient descent is the max and it is handled in
one of two ways in practice. Classical MIL algorithms treat the
label-determining instance as a latent variable and estimate it and
$\bbeta$ jointly with an EM-type procedure, such as EM-DD
\cite{zhang2002emdd}. However modern attention-based architectures instead use gradient descent and jointly learn the direction and temperature \cite{ilse2018attention,widrich2020deeprc}.  We therefore find it natural to ask: 
\begin{center}
How does gradient decent breakdown as the bag size and embedding dimension grow? 
\end{center}

We will write the learned parameter as $\bbeta=\rho\bb$, with $\bb$ constrained to the sphere $\|\bb\|^2=d$, and fit using the smooth approximation
\begin{equation}
  \hat w_\lambda(X\bb) = \sigma(\lambda X \bb)\cdot
(\rho X\bb) = \sum_{i=1}^n \rho y_i \left(\frac{e^{\lambda
  y_i}}{\sum_j e^{\lambda y_j}}\right),
  \label{eq:softmax_predictor}
\end{equation}
with $\sigma$ the row-wise softmax, $y_i=\langle\boldsymbol x_i,\bb\rangle$,
and $\lambda$ a \emph{learnable} inverse temperature. This interpolates between the hard max and a unweighted average as $\lambda$ tends towards $\infty$ and $0$ respectively. 
Gradient descent is then performed jointly in $\bb$, $\rho$, and $\lambda$, on the loss
\begin{equation}
L=\langle(w^k-\hat w_\lambda(X^k\bb))^2\rangle_\ell,
\end{equation}
where $\langle \cdot \rangle_\ell$ is the average over $\ell$ bags.
Evaluating the softmax on the normalized scores $X\bb$ ensures that
$\lambda$ measures the softmax sharpness independently of the learned
scale $\rho$.  We define 
$u=\langle\bb,\bb_{\rm true}\rangle/d$ as the normalized overlap between the
learned and true directions. 

\begin{figure}[t]
  \centering
  \includegraphics[width=\columnwidth]{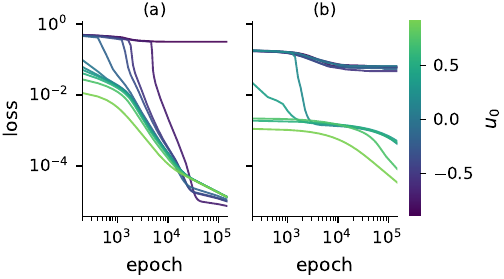}
  \caption{Training loss against epoch, for gradient descent from a grid of
    initial overlaps $u_0$ (colorbar) at two bag sizes.  (a) $d=22$, $n=3$ and (b) $d=150$,
    $n=1808$. Parameters are $(\lambda_0,\rho_0)=(0.2,1.7)$, $\alpha=0.05$,
    $\rho_{\rm true}=1.2$, $q=0.2$, $\ell=100$, with learning rate $0.1$. }
  \label{fig:loss_stall}
\end{figure}

Figure \ref{fig:loss_stall} illustrates that, as expected, the loss stalls as $d$ is increased with fixed $\alpha$. Moreover, depending on the initial overlap $u$, the trajectories have a qualitatively different convergence pattern, the signature of which is seen at smaller $d$ and $n$.

\textit{Loss asymptotics.---}%
We now derive the $d \to \infty$ limit of the normalized loss $\bar L :=
L/d^2$ with fixed $\alpha$.  The $d^2$ normalization is motivated by the fact that all asymptotic of $\hat w_\lambda$ scale with $d$, thus $L = O(d^2)$, as will be made clear from the calculations below.   Our asymptotic are based on a replacement of $\hat w_\lambda$ and $w_k$ with their almost sure limits. This approach allows us to explain some of the key phenomena exhibited by the trajectories in spite of the fact that  (1) the convergences breaks down at both phase transitions and (2) the convergence rate of $1/d$ makes it difficult to realize the limit numerically. We discuss this limitations and the precise nature of the convergence in the SI. 

Since $\bar L$ depends on $\bbeta$ only
through $(u,\rho)$, we write it as $\bar L(\lambda,u,\rho)$, and split it
according to whether the bag contains the planted instance,
\begin{equation}
  \bar L(\lambda,u,\rho) = q\,\bar L_+(\lambda,u,\rho) + (1-q)\,\bar L_-(\lambda,u,\rho).
\end{equation}

To evaluate $\bar L_+$ we need the large-$d$ limit of $\hat
w_\lambda(X\bbeta)$ on a positive bag, at overlap $u$ with the planted
pattern. Without loss of generality, let the first instance be the planted
one, $\boldsymbol x_1=\bbeta_{\rm true}$. The softmax weight of the planted instance is \begin{equation}
  \sigma_1(\bb) = \frac{e^{\lambda u\rho_{\rm true}d}}{e^{\lambda u\rho_{\rm true}d}+Z_n},
  \qquad
  Z_n = \sum_{i=2}^n e^{\lambda\langle\boldsymbol x_i,\bb\rangle},
  \label{eq:sigma_truth}
\end{equation}
using $\langle\boldsymbol x_1,\bb\rangle=u\rho_{\rm true}d$.  Eq.~\eqref{eq:sigma_truth}
shows that the behavior is governed by the competition between the
growth of $Z_n$ and the planted instance's weight,
$e^{\lambda u\rho_{\rm true}d}$.  To this end, we define \begin{equation}
 \phi_\alpha(\lambda):=  \lim_{d\to\infty}\frac{1}{\lambda d}\log Z_n. 
  \label{eq:noise_free_energy}
\end{equation}
Following \cite{lucibello2024exponential}, $\phi_{\alpha}$ can be computed exactly based on a mapping to the Random Energy Model (REM), a model of an equilibrium system  so disordered that the energies can be treated as independent random variables. The energies are modeled as iid Gaussian random variables, giving rise to a sum over log-normal random variables, exactly as in $Z_n$ (since the planted term has been removed). $\phi_\alpha(\lambda)$ is thus the scaled free energy of the REM. This free energy is known to exhibit a freezing transition between a regime where $Z_n \approx {\mathbb E}[e^{\lambda\langle\boldsymbol x_i,\bb\rangle}]$ and one where the sum is dominated by the extremal terms. 

Specifically, it can be shown that (see Supplemental Material Sec.~\ref{si:sec:gradient_flow})
\begin{equation}
  \phi_\alpha(\lambda) =
  \begin{cases}
    \alpha/\lambda+\lambda/2, & \lambda<\lambda^\star \quad (\mathrm H),\\
    \sqrt{2\alpha}, & \lambda\geq\lambda^\star \quad (\mathrm L),
  \end{cases}
  \label{eq:free_energy_piecewise}
\end{equation}
where $\lambda^\star=\sqrt{2\alpha}$ is the critical inverse temperature and $H$ and $L$ refer respectively to the high and low temperature phases. 

Comparing Eq.~\eqref{eq:noise_free_energy} to $u\rho_{\rm true}$ (the exponent of the planted instance) yields two phases for the limiting softmax weight corresponded to the retrieval and non-retrieval phase in \cite{lucibello2024exponential} \begin{equation}
  \lim_{d\to\infty}\sigma_1(\bb) =
  \begin{cases}
    1, & \rho_{\rm true}u > \phi_\alpha(\lambda) \quad (\mathrm R),\\
    0, & \rho_{\rm true}u < \phi_\alpha(\lambda) \quad (\mathrm N),
  \end{cases}
  \label{eq:retrieval_criterion}
\end{equation}
where R and N denote the retrieval and non-retrieval phases. In the
retrieval phase, all the softmax weight is on the planted instance, so
$\hat w_\lambda/d\to\langle\bbeta,\bbeta_{\rm true}\rangle/d=\rho\rho_{\rm
true}u$. In the non-retrieval phase, the high-temperature branch instead
behaves like an ordinary Gaussian average; using standard Gaussian integrals, we have
\begin{equation}
\frac{1}{d} \hat w_\lambda  \to \frac{{\mathbb E}[\langle x_i,\bbeta\rangle e^{\lambda \langle x_i,\bbeta\rangle}]}{{\mathbb E}[e^{\lambda \langle x_i,\bbeta\rangle}]} = \lambda\rho. 
\end{equation}
 The low-temperature branch is instead
dominated by the single largest $\langle x_i,\bbeta\rangle$ term, which we have already argued grows as $\rho\sqrt{2\alpha}\,d$,
giving $\hat w_\lambda/d\to\sqrt{2\alpha}\,\rho$. Negative bags contain no
planted instance and therefore admit only the two non-retrieval forms. Writing
$\bar w^{\pm}:=\lim_{d\to\infty}\hat w_\lambda(X^{\pm}\bbeta)/d$ for the
resulting positive/negative-bag output, the H/L and R/N transitions are
independent and combine into the four phases of Table~\ref{tab:phases}.  In each phase, the conditional loss can be
written as the square of a corresponding residual, $\Delta$. 

\begin{table}[t]
\centering
 \begin{tabular}{r|l|l|l|l}
  Phase & $\bar w^+$ & $\bar w^-$ & $\Delta_+$ & $\Delta_-$ \\
  \hline\hline
  HR & \cellcolor{blue!12}$\rho\rho_{\rm true}u$ & \cellcolor{green!15}$\lambda\rho$ & \cellcolor{orange!15}$\rho_{\rm true}^2-\rho\rho_{\rm true}u$ &
  \cellcolor{purple!12}$\sqrt{2\alpha}\rho_{\rm true}-\lambda\rho$ \\
  LR & \cellcolor{blue!12}$\rho\rho_{\rm true}u$ & \cellcolor{cyan!15}$\sqrt{2\alpha}\,\rho$ & \cellcolor{orange!15}$\rho_{\rm true}^2-\rho\rho_{\rm true}u$ &
  \cellcolor{yellow!20}$\sqrt{2\alpha}(\rho_{\rm true}-\rho)$ \\
  HN & \cellcolor{green!15}$\lambda\rho$ & \cellcolor{green!15}$\lambda\rho$ & $\rho_{\rm true}^2-\lambda\rho$ & \cellcolor{purple!12}$\sqrt{2\alpha}\rho_{\rm true}-\lambda\rho$ \\
  LN & \cellcolor{cyan!15}$\sqrt{2\alpha}\,\rho$ & \cellcolor{cyan!15}$\sqrt{2\alpha}\,\rho$ & $\rho_{\rm true}^2-\sqrt{2\alpha}\rho$ & \cellcolor{yellow!20}$\sqrt{2\alpha}(\rho_{\rm true}-\rho)$ \\
  \hline
 \text{Exact} & $\rho_{\rm true}^2$ & $\sqrt{2\alpha}\,\rho_{\rm true}$ & &
  \end{tabular}
\caption{Asymptotic normalized bag output $\hat w_\lambda(X^{\pm}\bbeta)/d$
  in each of the four phases (H/L $\times$ R/N), the corresponding
  large-$d$ limit of the target $w^k/d$, which is set by the true
  generative process alone and so is identical across all four phases,
  and the resulting per-bag-type residuals $\Delta_\pm$. The color of the cells is used to highlight which quantities are the same.}
\label{tab:phases}
\end{table}

\textit{Gradient flow ODEs.---}%
The continuous gradient flow is
\begin{equation}
  \dot\lambda=-\partial_\lambda\bar L, \qquad
  \dot\rho=-\partial_\rho\bar L, \qquad
  \dot{\bb}=-P_{\bb}\nabla_{\bb}\bar L,
  \label{eq:gradient_flow_def}
\end{equation}
where $P_{\bb}v:=v-\langle v,\bb\rangle\bb/d$ projects onto the tangent
space of the sphere $\sqrt d\,S^{d-1}$ at $\bb$. Using that $P_{\bb}$ is
self-adjoint,
\begin{equation}
  \dot u = \frac1d\langle\dot{\bb},\bb_{\rm true}\rangle = -\frac1d\langle\nabla_{\bb}\bar L,\,P_{\bb}\bb_{\rm true}\rangle.
\end{equation}
$\bar L$ depends on $\bb$ only through $u$, so $\nabla_{\bb}\bar
L=\bb_{\rm true}\partial_u\bar L$. Using $\langle\bb_{\rm
true},P_{\bb}\bb_{\rm true}\rangle=d(1-u^2)$ then gives
$\dot u=-(1-u^2)\partial_u\bar L$.

For $c\in\{\lambda,u,\rho\}$ the partial derivatives of $\bar L$ are approximated as 
\begin{equation}
  -\partial_c\bar L \approx 2\Big[q\,\Delta\,\partial_c\bar w^+ + (1-q)\,\Delta\,\partial_c\bar w^-\Big],
  \label{eq:master_gradient_rule}
\end{equation}
with $\Delta$ the residual from Table~\ref{tab:phases}
relevant to that phase and bag type, resulting in a system of piecewise smooth  ODEs. The complete set of equations are given in  Supplemental Material Sec.~\ref{si:sec:gradient_flow}. 

We now consider the fate of a trajectory beginning in the H phase. For a trajectory initially in the HN phase, $\dot u=0$, resulting in a
two-dimensional system for $\lambda$ and $\rho$:
\begin{align}
\dot \lambda &= 2\rho\big[q\rho_{\rm true}^2+(1-q)\sqrt{2\alpha}\rho_{\rm true}-\lambda\rho\big]
\end{align}
and $\dot\rho=(\lambda/\rho)\dot\lambda$. Note that since $d( \rho^2)/dt = 2 \lambda \dot \lambda$, $C = \lambda_0^2 - \rho_0^2$ is conserved along trajectories.  The positive fixed points are the hyperbola
\begin{equation}\label{eq:ss1}
  \lambda\rho = q\rho_{\rm true}^2+(1-q)\sqrt{2\alpha}\rho_{\rm true},
\end{equation}
restricted to the HN region. 
If $\lambda_{\rm fp}<\sqrt{2\alpha}$, the trajectory equilibrates in HN
with $u$ still fixed at its initial value. If
$\lambda_{\rm fp}>\sqrt{2\alpha}$, it reaches the temperature boundary
first and enters LN, where $\dot \lambda = \dot u =0$. 

If the trajectory starts in HR instead, all three variables evolve:
\begin{align}
\dot\lambda &= 2(1-q)\rho(\sqrt{2\alpha}\rho_{\rm true}-\lambda\rho),\\
\dot u &= 2q\rho\rho_{\rm true}^2(\rho_{\rm true}-\rho u)(1-u^2),\\
\dot\rho &= 2q u\rho_{\rm true}^2(\rho_{\rm true}-\rho u)\notag\\
 &\quad\quad +2(1-q)\lambda(\sqrt{2\alpha}\rho_{\rm true}-\lambda\rho).
\end{align}
For an interior fixed point, the two residuals must vanish:
\begin{equation}\label{eq:ss2}
  \rho u=\rho_{\rm true},\qquad
  \lambda\rho=\sqrt{2\alpha}\rho_{\rm true}.
\end{equation}
Thus the HR fixed set is the curve
$\lambda=\sqrt{2\alpha}\,u$, $\rho=\rho_{\rm true}/u$, restricted to the
part lying inside HR.

\begin{figure}[t]
  \centering
  \includegraphics[width=\columnwidth,trim=0 0 0 11bp,clip]{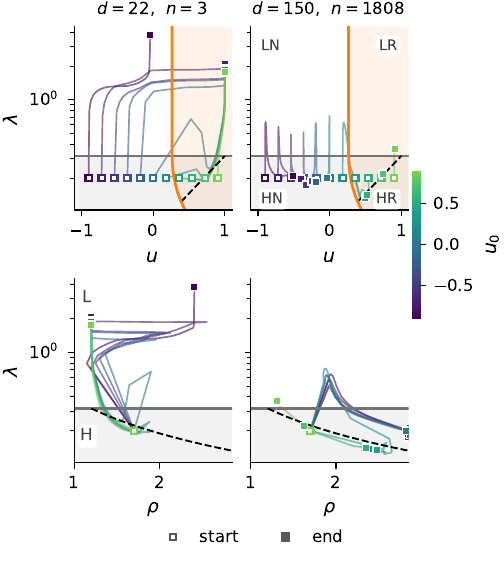}
  \caption{Gradient-descent trajectories with different $u_0$ (colorbar). The top row shows the dynamics in $(u,\lambda)$ space with the H/L and R/N phase boundaries shown. The bottom panels show trajectories in $(\rho,\lambda)$ space, with the H/L boundary shows. the dashed lines denote the invariant manifold defined by Eqs. \ref{eq:ss1} and \ref{eq:ss2}. Parameters are $(\lambda_0,\rho_0)=(0.2,1.7)$, $\alpha=0.05$,
    $\rho_{\rm true}=1.2$, $q=0.2$, $\ell=100$, run to flow time $t=15000$ (learning rate $0.1$).}
  \label{fig:phase_transition}
\end{figure}

\textit{Numerical results.---} This analysis suggests that as $d \to \infty$ the gradient flow exhibits two qualitatively distinct failure modes depending on whether they begin in the R vs N phase. Figure~\ref{fig:loss_stall} illustrates that this behavior by plotting the loss as a function of time (the epoch) gradient descent for different $u_0$. In this numerical experiments, we are able to see the phase transition is present even at moderate size. In particular, in panel (a) where $d=22$, it is see that trajectories exhibit qualitatively distinct dynamics depending on whether they begin in R or N phase. In panel (b) $d$ is increased and our simulations also show that for fixed $\alpha$ the loss plateaus for all initial $u$, consistent with the theory.

To further illustrate that the limit ODEs indeed capture the dynamics of gradient descent for a trajectory beginning in $H$, we show the trajectories themselves in  Figure~\ref{fig:phase_transition} in the $\lambda$-$u$ and $\lambda$-$\rho$ planes. The invariant manifolds from Eqs. \ref{eq:ss1} and \ref{eq:ss2} are shown as dashed lines. As predicted by our theory, as $d$ grows, we can clearly see the transition between the HN and HR initial conditions. Surprisingly, even though the limit ODEs are a crude approximation, the fixed lines appear to explain much of the variation in the long-time behavior for the trajectories beginning in HN. 

\begin{figure*}[t]
  \centering
  \includegraphics[width=\textwidth]{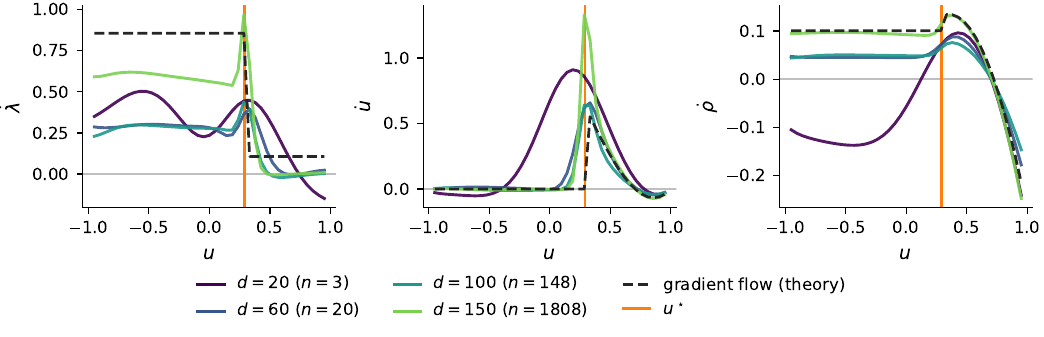}
  \caption{Convergence of the finite-size training gradient to the
    asymptotic gradient-flow field. Empirical
    $(\dot\lambda,\dot u,\dot\rho)$ compared to the $d \to \infty$ theory (dashed line). Parameters are  $(\lambda_0,\rho_0)=(0.2,1.7)$, $\alpha=0.05$, $\rho_{\rm true}=1.2$,
    $q=0.2$, $\ell=100$. The vertical line marks $u^\star$. }
  \label{fig:gradient_curves}
\end{figure*}

In the SI, we show that away from the phase transition, the error in the ODEs scales as $O(\ln d /d)$; however, the pre-factor diverges as we approach the phase transition.  As more direct comparisons of the asymptotic theory to the gradient
itself, we have shown the  initial gradient as a function $u$ in Figure~\ref{fig:gradient_curves}. In this figure, we can see that the limit ODEs predict the gradients with respect to $\rho$ and $u$ (away from the phase boundary). The gradient with respect to $\lambda$ is qualitatively similar, being a step function (away from the phase boundary). 

\textit{Discussion.---}%
We have studied learning dynamics in a minimal model of multiple instance learning, where there is an extremal selection rule for a planted instance. This problem is meant to exemplify the challenges of applying gradient descent to models where the strength of the nonlinearity (the temperature) is learned, and learning naturally happens into regimes of vanishing gradients. Using tools from disordered systems -- namely the random energy model  -- and applying previous results on Hopfield networks, we have derived limiting ODEs for the low dimensional dynamics of the temperature and regression coefficients. Our analysis, which accurately predicts numerical experiments with Gaussian data, suggests that one expects to find abrupt transitions in the speed of learning for this system depending on the misalignment of the initial regression coefficients. This suggests it would be worthwhile exploring other methods besides gradient descent for more efficiently tuning the regression coefficients during the early stages of learning. For example, coupling gradient descent with EM-based approaches \cite{Levien2026EMQueryValue}. 

\textit{Code availability.---} The code is available at \url{https://github.com/elevien/extremal-mil}.

\begin{acknowledgments}
E.L.\ and T.G.\ acknowledge support from the National Institute for Theory and
Mathematics in Biology through the National Science Foundation (grant number
DMS-2235451) and the Simons Foundation (grant number MP-TMPS-00005320).
E.L.\ acknowledges support from the National Science Foundation (grant number
DMS-2527337) and a Dartmouth College Burke Research Initiation Award.
\end{acknowledgments}

\bibliographystyle{apsrev4-2}
\bibliography{mil_prl}

@article{dietterich1997solving,
  author  = {Dietterich, Thomas G. and Lathrop, Richard H. and Lozano-P{\'e}rez, Tom{\'a}s},
  title   = {Solving the multiple instance problem with axis-parallel rectangles},
  journal = {Artificial Intelligence},
  volume  = {89},
  number  = {1--2},
  pages   = {31--71},
  year    = {1997}
}

@book{bovier2006statistical,
  title={Statistical mechanics of disordered systems: a mathematical perspective},
  author={Bovier, Anton},
  volume={18},
  year={2006},
  publisher={Cambridge University Press}
}

@inproceedings{maron1998framework,
  author    = {Maron, Oded and Lozano-P{\'e}rez, Tom{\'a}s},
  title     = {A framework for multiple-instance learning},
  booktitle = {Advances in Neural Information Processing Systems},
  volume    = {10},
  year      = {1998},
  publisher = {MIT Press}
}

@inproceedings{ilse2018attention,
  author    = {Ilse, Maximilian and Tomczak, Jakub M. and Welling, Max},
  title     = {Attention-based deep multiple instance learning},
  booktitle = {Proceedings of the 35th International Conference on Machine Learning},
  pages     = {2127--2136},
  year      = {2018},
  publisher = {PMLR}
}

@inproceedings{widrich2020deeprc,
  author    = {Widrich, Michael and Sch{\"a}fl, Bernhard and Pavlovi{\'c}, Milena and
               Ramsauer, Hubert and Gruber, Lukas and Holzleitner, Markus and
               Brandstetter, Johannes and Sandve, Geir Kjetil and Greiff, Victor and
               Hochreiter, Sepp and Klambauer, G{\"u}nter},
  title     = {Modern {H}opfield Networks and Attention for Immune Repertoire Classification},
  booktitle = {Advances in Neural Information Processing Systems},
  volume    = {33},
  pages     = {18832--18845},
  year      = {2020}
}

@inproceedings{zhang2002emdd,
  author    = {Zhang, Qi and Goldman, Sally A.},
  title     = {{EM-DD}: An Improved Multiple-Instance Learning Technique},
  booktitle = {Advances in Neural Information Processing Systems},
  volume    = {15},
  year      = {2002},
  publisher = {MIT Press}
}

@article{ben2005limit,
  author  = {Ben Arous, G{\'e}rard and Bogachev, Leonid V and Molchanov, Stanislav A},
  title   = {Limit theorems for sums of random exponentials},
  journal = {Probability Theory and Related Fields},
  volume  = {132},
  pages   = {579--612},
  year    = {2005}
}

@article{hopfield1982neural,
  author  = {Hopfield, John J},
  title   = {Neural networks and physical systems with emergent collective computational abilities},
  journal = {Proceedings of the National Academy of Sciences},
  volume  = {79},
  number  = {8},
  pages   = {2554--2558},
  year    = {1982}
}

@article{lucibello2024exponential,
  author  = {Lucibello, Carlo and M{\'e}zard, Marc},
  title   = {Exponential capacity of dense associative memories},
  journal = {Physical Review Letters},
  volume  = {132},
  number  = {7},
  pages   = {077301},
  year    = {2024}
}

@book{mezard2009information,
  author    = {Mezard, Marc and Montanari, Andrea},
  title     = {Information, Physics, and Computation},
  publisher = {Oxford University Press},
  year      = {2009}
}

@article{ramsauer2020hopfield,
  author  = {Ramsauer, Hubert and Sch{\"a}fl, Bernhard and Lehner, Johannes and Seidl, Philipp
             and Widrich, Michael and Adler, Thomas and Gruber, Lukas and Holzleitner, Markus
             and Pavlovi{\'c}, Milena and Sandve, Geir Kjetil and others},
  title   = {Hopfield networks is all you need},
  journal = {arXiv preprint arXiv:2008.02217},
  year    = {2020}
}

@article{Carbonneau2018survey,
  author  = {Carbonneau, Marc-Andr{\'e} and Cheplygina, Veronika and Granger, Eric and Gagnon, Ghyslain},
  title   = {Multiple instance learning: A survey of problem characteristics and applications},
  journal = {Pattern Recognition},
  volume  = {77},
  pages   = {329--353},
  year    = {2018},
  publisher = {Elsevier}
}

@inproceedings{maron1998learning,
  author    = {Maron, Oded},
  title     = {Learning from ambiguity},
  booktitle = {Advances in Neural Information Processing Systems},
  volume    = {11},
  year      = {1998}
}

@article{vuckovic2020mathematical,
  author  = {Vuckovic, James and Baratin, Aristide and Tachet des Combes, Remi},
  title   = {A mathematical theory of attention},
  journal = {arXiv preprint arXiv:2007.02876},
  year    = {2020}
}

@inproceedings{RayPage2001MIR,
  author    = {Soumya Ray and David Page},
  title     = {Multiple‐Instance Regression},
  booktitle = {Proceedings of the International Conference on Machine Learning (ICML)},
  year      = {2001},
  pages     = {425--432},
}

@article{WagstaffLane2007StructuredMIR,
  author    = {Kiri L. Wagstaff and Terran Lane},
  title     = {Multiple‐Instance Regression with Structured Data},
  journal   = {Proc. of International Conference on Machine Learning (ICML) Workshop},
  year      = {2007},
}

@article{dooly2002multiple,
  title={Multiple-instance learning of real-valued data},
  author={Dooly, Daniel R and Zhang, Qi and Goldman, Sally A and Amar, Robert A},
  journal={Journal of Machine Learning Research},
  volume={3},
  pages={651--678},
  year={2002}
}

@article{dooly2006real,
  title={Real-valued multiple-instance learning with queries},
  author={Dooly, Daniel R and Goldman, Sally A and Kwek, Stephen S},
  journal={Journal of Computer and System Sciences},
  volume={72},
  number={1},
  pages={1--15},
  year={2006},
  publisher={Elsevier}
}

@inproceedings{Chauhan2024GenLearnMIR,
  author    = {Kushal Chauhan and Rishi Saket and Lorne Applebaum and Ashwinkumar Badanidiyuru and Chandan Giri and Aravindan Raghuveer},
  title     = {Generalization and Learnability in Multiple Instance Regression},
  booktitle = {Proceedings of Uncertainty in Artificial Intelligence (UAI)},
  year      = {2024},
}

@inproceedings{Andrews2002MISVM,
  author    = {Andrews, Stuart and Tsochantaridis, Ioannis and Hofmann, Thomas},
  title     = {Support Vector Machines for Multiple-Instance Learning},
  booktitle = {Advances in Neural Information Processing Systems},
  volume    = {15},
  pages     = {561--568},
  year      = {2002}
}

@inproceedings{Gartner2002MIKernels,
  author    = {G{\"a}rtner, Thomas and Flach, Peter A. and Kowalczyk, Adam and Smola, Alex J.},
  title     = {Multi-Instance Kernels},
  booktitle = {Proceedings of the Nineteenth International Conference on Machine Learning},
  pages     = {179--186},
  year      = {2002}
}

@article{diochnos2012multiple,
  title={On multiple-instance learning of halfspaces},
  author={Diochnos, Dimitrios I and Sloan, Robert H and Tur{\'a}n, Gy},
  journal={Information Processing Letters},
  volume={112},
  number={23},
  pages={933--936},
  year={2012},
  publisher={Elsevier}
}

@article{doran2016multiple,
  title={Multiple-instance learning from distributions},
  author={Doran, Gary and Ray, Soumya},
  journal={Journal of Machine Learning Research},
  volume={17},
  number={128},
  pages={1--50},
  year={2016}
}

@article{sliwoski2014computational,
  title={Computational methods in drug discovery},
  author={Sliwoski, Gregory and Kothiwale, Sandeepkumar and Meiler, Jens and Lowe Jr, Edward W},
  journal={Pharmacological Reviews},
  volume={66},
  number={1},
  pages={334--395},
  year={2014},
  publisher={Elsevier}
}

@article{jain1994compass,
  title={Compass: predicting biological activities from molecular surface properties. Performance comparisons on a steroid benchmark},
  author={Jain, Ajay N and Koile, Kimberle and Chapman, David},
  journal={Journal of Medicinal Chemistry},
  volume={37},
  number={15},
  pages={2315--2327},
  year={1994},
  publisher={ACS Publications}
}

@article{zankov2024chemical,
  title={Chemical complexity challenge: Is multi-instance machine learning a solution?},
  author={Zankov, Dmitry and Madzhidov, Timur and Varnek, Alexandre and Polishchuk, Pavel},
  journal={Wiley Interdisciplinary Reviews: Computational Molecular Science},
  volume={14},
  number={1},
  pages={e1698},
  year={2024},
  publisher={Wiley Online Library}
}

@inproceedings{Vaswani2017,
  author    = {Vaswani, Ashish and Shazeer, Noam and Parmar, Niki and Uszkoreit, Jakob and Jones, Llion and Gomez, Aidan N. and Kaiser, {\L}ukasz and Polosukhin, Illia},
  title     = {Attention Is All You Need},
  booktitle = {Advances in Neural Information Processing Systems},
  volume    = {30},
  year      = {2017},
  publisher = {Curran Associates, Inc.}
}

@inproceedings{szabo2015two,
  title={Two-stage sampled learning theory on distributions},
  author={Szab{\'o}, Zolt{\'a}n and Gretton, Arthur and P{\'o}czos, Barnab{\'a}s and Sriperumbudur, Bharath},
  booktitle={Artificial Intelligence and Statistics},
  pages={948--957},
  year={2015},
  organization={PMLR}
}

@article{derrida1980random,
  author  = {Derrida, Bernard},
  title   = {Random-Energy Model: Limit of a Family of Disordered Models},
  journal = {Physical Review Letters},
  volume  = {45},
  number  = {2},
  pages   = {79--82},
  year    = {1980}
}

@article{derrida1981random,
  author  = {Derrida, Bernard},
  title   = {Random-energy model: An exactly solvable model of disordered systems},
  journal = {Physical Review B},
  volume  = {24},
  number  = {5},
  pages   = {2613--2626},
  year    = {1981}
}

@article{soudry2018implicit,
  author  = {Soudry, Daniel and Hoffer, Elad and Nacson, Mor Shpigel and Gunasekar, Suriya and Srebro, Nathan},
  title   = {The Implicit Bias of Gradient Descent on Separable Data},
  journal = {Journal of Machine Learning Research},
  volume  = {19},
  number  = {70},
  pages   = {1--57},
  year    = {2018},
  note    = {arXiv:1710.10345}
}

@book{leadbetter1983extremes,
  title={Extremes and Related Properties of Random Sequences and Processes},
  author={Leadbetter, M. R. and Lindgren, G. and Rootz{\'e}n, H.},
  publisher={Springer},
  address={New York},
  year={1983}
}

@article{Levien2026EMQueryValue,
  author  = {Levien, Ethan},
  title   = {{EM}-based iterations for multiple instance learning on a query-value model},
  journal = {arXiv preprint arXiv:2607.17404},
  year    = {2026}
}

\clearpage
\onecolumngrid
\let\addcontentsline\origaddcontentsline

\setcounter{secnumdepth}{3}
\setcounter{section}{0}
\setcounter{equation}{0}
\setcounter{figure}{0}
\setcounter{table}{0}
\renewcommand{\thesection}{S\arabic{section}}
\renewcommand{\theequation}{S\arabic{equation}}
\renewcommand{\thefigure}{S\arabic{figure}}
\renewcommand{\thetable}{S\arabic{table}}
\renewcommand{\theHsection}{S\arabic{section}}
\renewcommand{\theHequation}{S\arabic{equation}}
\renewcommand{\theHfigure}{S\arabic{figure}}
\renewcommand{\theHtable}{S\arabic{table}}

\begin{center}
{\large\bfseries Supplemental Material}
\end{center}

\tableofcontents

\section{Background: REM and EVT}
\label{si:sec:asymptotics}

In this section we present background on the Random Energy Model and extremal value statistics. 

\subsection{Extreme value statistics}
Let $x_1,\dots,x_n$ be i.i.d.\ ${\rm Normal}(0,s^2)$ and let
$M_n=\max_{i\le n}x_i$. A classical result of extreme value statistics is the following \cite{leadbetter1983extremes}.
\begin{theorem}[Almost sure growth of the maximum]
  \label{si:thm:as_max}
  As $n\to\infty$,
  \begin{equation}
    \frac{M_n}{s\sqrt{2\ln n}}\to1 \quad\text{almost surely.}
  \end{equation}
\end{theorem}
 The
generalized extreme value theorem (Fisher--Tippett--Gnedenko) describes the
fluctuations around it \cite{leadbetter1983extremes}.
\begin{theorem}[Generalized extreme value theorem]
  \label{si:thm:gev}
  As $n\to\infty$, 
  \begin{equation}
 {\mathbb P}((M_n-b_n)/a_n  >m) \to 1-e^{-e^{-m}},
  \end{equation}
where 
  \begin{equation}
    b_n=s\left(\sqrt{2\ln n}-\frac{\ln\ln n+\ln4\pi}{2\sqrt{2\ln n}}\right),
    \qquad a_n=\frac{s}{\sqrt{2\ln n}}.
  \end{equation}
That is, $(M_n-b_n)/a_n$ converges to a Gumbel random variable. 
\end{theorem}
The fluctuations of $M_n$ have width $a_n=O(s/\sqrt{\ln n})$,
which is small compared to the location $b_n=O(s\sqrt{\ln n})$. However, the residual $O(\ln\ln n/\ln n)$, so the approach to
the almost sure limit of Theorem~\ref{si:thm:as_max} is slow. In our setting
$n=e^{\alpha d}$ and the variance $s^2=d$ grows with $n$. Writing the
scores as $\sqrt d$ times standard normals and applying the theorems with
$s=1$ gives $M_n\approx d\sqrt{2\alpha}$
with Gumbel fluctuations of order one and a relative correction of order
$\ln d/d$.

\subsection{The random energy model}
\label{si:sec:rem}

Random energy models (REM)  \cite{derrida1980random,derrida1981random} are toy models of highly disordered systems -- ones where the energy landscape has many local minima. To study generic properties of such systems, the idea is to randomly sample the energies. This so-called \emph{quenched} disorder in the energy landscape is another level of randomness on top of the statistical ensemble itself. These two forms of randomness correspond to the distribution of bags and the within-bag distribution, respectively.

We consider the simplest REM, where the energies are iid Normal.
The partition function for this model takes the form 
\begin{equation}
  Z_{{\rm REM},N} = \sum_{i=1}^{2^N}e^{-\beta\sqrt N x_i}, \qquad x_i\sim{\rm Normal}(0,1). 
\end{equation}
This model has the free energy \cite{derrida1980random} 
\begin{equation}\label{eq:REMfree}
 \bar{\phi}_{{\rm REM}}(\beta) :=  -\lim_{N\to\infty}\frac{1}{N\beta}\ln Z_{{\rm REM},N} =
  \begin{cases}
    -\beta/2-\beta_c^2/(2\beta), & \beta\leq\beta_c, \\
    -\beta_c, & \beta>\beta_c,
  \end{cases}
\end{equation}
with $\beta_c=\sqrt{2\ln2}$. The transition at $\beta_c$ separates a
high-temperature (H) phase in which the sum is dominated by the exponentially many
configurations, from a low-temperature phase (L) in
which it is dominated by the single lowest-energy configuration. 

One can derive the transition in Eq.~\eqref{eq:REMfree} as the point where the average of $Z_{{\rm REM},N}$ grows as fast as the maximum weight. By Theorem~\ref{si:thm:as_max}, $\max_i(-x_i)\approx\sqrt{2N\ln2}$, so this happens when
\begin{equation}
\mathbb E[Z_{{\rm REM},N}] = e^{N(\ln2+\beta^2/2)} = e^{\beta N\sqrt{2\ln2}} \approx \max_i e^{-\beta\sqrt N x_i},
\end{equation}
that is, when $\beta=\sqrt{2\ln2}=\beta_c$.

Another important quantity in REM theory, which appears in our analysis below, is the \emph{replica overlap}. Let $\sigma_i=e^{-\beta\sqrt Nx_i}/Z_{{\rm REM},N}$ be the Gibbs weights
of $Z_{{\rm REM},N}$, and define the overlap
\begin{equation}
  \psi_N(\beta)=\sum_{i=1}^{2^N}\sigma_i^2.
  \label{si:eq:overlap}
\end{equation}
$\psi_N(\beta)$ is the probability that two independent draws from the
Gibbs weights pick the same configuration, i.e.\ the mass at $1$ of the
replica overlap distribution.
It is useful that the derivative of $\ln Z_{{\rm REM},N}$ with respect to the inverse
temperature is related to the
overlap by the following Lemma,  Gaussian integration by parts.
\begin{lemma}[{\cite[Proposition 9.5.1]{bovier2006statistical}}]
  \label{si:lem:overlap}
  For every $N$ and $\beta$, 
  \begin{equation}
    \frac1N\mathbb E\big[\partial_\beta\ln Z_{{\rm REM},N}\big]=\beta\Big(1-\mathbb E\big[\psi_N(\beta)\big]\Big).
  \end{equation}
\end{lemma}

The limit of the overlap is known for the REM.
\begin{theorem}[]
  \label{si:thm:overlap}
  \begin{equation}
    \lim_{N\to\infty}\mathbb E\big[\psi_N(\beta)\big]=
    \begin{cases}
      0, & \beta\le\beta_c,\\
      1-\beta_c/\beta, & \beta>\beta_c.
    \end{cases}
  \end{equation}
\end{theorem}

\subsection{Weak self-averaging transition}
Subsequent work
\cite{ben2005limit} describes the limiting law of $Z_{{\rm REM},N}$ in
more detail. This is closely related to, and in fact can be viewed as a special case of, classical limit theorems for triangular arrays. The limit laws of such sums are $\alpha$-stable distributions. The simplest case is that of a power-law sum, where the generalized central limit theorem describes two transitions: one where the CLT breaks down but the LLN still holds and the fluctuations scale as $N^{1/\alpha}$, and one where the LLN breaks down, which corresponds to the freezing transition in the REM. The second phase transition is derived by comparing the growth of the second moment to the growth of the max. The sum of the second moments of the weights grows as
\begin{equation}
\sum_{i=1}^{2^N}\mathbb E\big[e^{-2\beta\sqrt N x_i}\big] = e^{N(\ln2+2\beta^2)},
\end{equation}
while the square of the maximum weight grows as $e^{2\beta N\sqrt{2\ln2}}$. These rates coincide when $\beta=\sqrt{\ln2/2}=\beta_c/2$. 

\section{Gradient Flow Equations}
\label{si:sec:gradient_flow}
\subsection{Setup and normalization}

Recall that we represent a bag as a matrix 
$X_k\in\mathbb R^{n\times d}$ and sample the entries from iid standard normal and then replacing a row with 
$\bbeta_{\rm true}$ with probability $q$, yielding a ``positive'' bag.
We fix $\bbeta_{\rm true}=\sqrt d\,\rho_{\rm true}e_1$ without loss of
generality, where $e_1$ is the first coordinate vector, and parameterize
the learned direction as
$\bb=\sqrt d\,(u e_1+\sqrt{1-u^2}\,\boldsymbol\delta)$, with
$\boldsymbol\delta\perp e_1$, $\|\boldsymbol\delta\|=1$, and
$\bbeta=\rho\bb$. The loss
\begin{equation}
  \bar L = \frac{1}{d^2}\,\mathbb E\big[(w^k-\hat w_\lambda(X^k\bbeta))^2\big]
\end{equation}
We consider each of the terms $\mathbb E\big[(w^k)^2]$, $\mathbb E\big[w^k\hat w_\lambda(X^k\bbeta)\big]$ and $\mathbb E\big[(\hat w_\lambda(X^k\bbeta))^2\big]$.

\subsubsection{True output second moment}
By Theorem~\ref{si:thm:gev}, the max over background instances has the limit 
\begin{equation}
d\left(\frac{m_n}{\rho_{\rm true}}-\sqrt{2\alpha}+\frac{\ln(\alpha d)+\ln4\pi}{2\sqrt{2\alpha}\,d}\right)\xrightarrow{d}\frac{G}{\sqrt{2\alpha}},
  \qquad
  m_n:=\frac{1}{d}{\rm max}_{i >1}\,  {\boldsymbol x}_i^k \cdot \bbeta_{\rm true},
  \label{si:eq:Mtrue_expansion}
\end{equation}
where $G$ is a standard Gumbel random variable defined by Theorem \ref{si:thm:gev}. 

In the identifiable regime
$\alpha<\rho_{\rm true}^2/2$, Eq.~\eqref{si:eq:Mtrue_expansion} gives
\begin{equation}
\frac{1}{d}{\rm max}_{i >1}\,  {\boldsymbol x}_i^k \cdot \bbeta_{\rm true} \to\rho_{\rm true}\sqrt{2\alpha}<\rho_{\rm true}^2
\end{equation}
 in
probability, so the
planted instance realizes the maximum of a positive bag with probability
tending to one. Hence $w^k/d\sim \rho_{\rm true}^2$ on positive bags and
$w^k/d \sim  m_n$ on negative bags. We write $\mathbb E^{\pm}$ to indicate the expectation over positive (respectively negative) bags. 

We then have 
\begin{align}  \label{si:eq:target_second_moment}
\frac{1}{d^2}\mathbb E^{+}[(w^k)^2] \approx \rho_{\rm true}^4,\quad 
\frac{1}{d^2}\mathbb E^{-}[(w^k)^2] \approx \rho_{\rm true}^2 2 \alpha - \frac{ \rho_{\rm true}^2}{d} \left(\ln (\alpha d) + \ln 4 \pi - 2 \gamma_E\right) 
\end{align}
where $\gamma_E$ is the Euler–Mascheroni constant, which comes from a standard result about the Gumbel distribution. 

\subsubsection{Model output}
Following the main text, we define the auxiliary partition function
\begin{equation}
  Z_n=\sum_{i=2}^n e^{\lambda\,{\boldsymbol x}_i^k\cdot\bb},
\end{equation}
where on a positive bag ${\boldsymbol x}_1^k=\bbeta_{\rm true}$ is the
planted instance. Using ${\boldsymbol x}_1^k\cdot\bb=u\rho_{\rm true}d$, the
softmax weight on the planted instance is
\begin{equation}  \label{si:eq:sigma_truth}
  \sigma_1(\bb) = \frac{e^{\lambda u\rho_{\rm true}d}}{e^{\lambda u\rho_{\rm true}d}+Z_n}
\end{equation}
The model output is then
\begin{equation}
  \frac1d\hat w_\lambda(X^k\bbeta)=\sigma_1\,\rho\rho_{\rm true}u+(1-\sigma_1)\,\frac{\rho}{d}\,\partial_\lambda\ln Z_n,
  \label{si:eq:model_output_Zn}
\end{equation}
On negative bags, ${\boldsymbol x}_1^k$ is a background instance and the
output is
\begin{equation}
  \frac1d\hat w_\lambda(X^k\bbeta)=\frac{\rho}{d}\,\partial_\lambda\ln Z_{n+1}
\end{equation}

 The limiting behavior comes from mapping $Z_n$ to $Z_{{\rm REM},N}$. Matching the
number of terms, $2^N=e^{\alpha d}$, and the exponents,
$\beta\sqrt N x_i=\lambda\,{\boldsymbol x}_i^k\cdot\bb$ in distribution,
gives the correspondence to the REM in Table~\ref{si:tab:rem_dictionary}. 

\begin{table}[h]
\centering
\begin{tabular}{l|l|l}
  & REM & MIL Model \\
  \hline\hline
  System size & $N$ & $\alpha d/\ln2$ \\
  Number of terms & $2^N$ & $n$ \\
  Exponents & $-\beta\sqrt N\,x_i$ & $\lambda\,{\boldsymbol x}_i^k\cdot\bb$ \\
  Inverse temperature & $\beta$ & $\lambda\sqrt{\ln2/\alpha}$ \\
  Freezing transition & $\beta_c=\sqrt{2\ln2}$ & $\lambda^\star=\sqrt{2\alpha}$ \\
  CLT transition & $\beta_c/2$ & $\lambda^\star/2$ \\
  Partition function & $Z_{{\rm REM},N}$ & $Z_n$
\end{tabular}
\caption{Correspondence between the REM of Sec.~\ref{si:sec:rem} and the
  off-target partition function $Z_n$. The sign of the exponents is
  immaterial because their distribution is symmetric.}
\label{si:tab:rem_dictionary}
\end{table}

With $\sigma_i$ the softmax weights of the off-target instances,
Lemma~\ref{si:lem:overlap} under this correspondence gives
\begin{equation}
  \frac1d\mathbb E\big[\partial_\lambda\ln Z_n\big]=\lambda\Big(1-\mathbb E\Big[\sum_{i=2}^n\sigma_i^2\Big]\Big)
  \to
  \begin{cases}
    \lambda, & \lambda<\lambda^\star\ ({\rm H}),\\
    \lambda^\star=\sqrt{2\alpha}, & \lambda>\lambda^\star\ ({\rm L}),
  \end{cases}
\end{equation}
where the limit follows from Theorem~\ref{si:thm:overlap}. For $\lambda<\lambda^\star$ the
softmax average is spread over exponentially many instances, and the
fluctuations of $\frac1d\partial_\lambda\ln Z_n$ are exponentially small.
Note that this holds on both sides of the second transition at $\lambda^\star/2$,
which changes the fluctuations of $Z_n$ but not those of the output. 

For
$\lambda>\lambda^\star$, $Z_n$ is dominated by a few instances near the
maximum, and the finer expansion is 
\begin{equation}
  \frac1d\partial_\lambda\ln Z_n=\sqrt{2\alpha}-\frac{\ln(\alpha d)+\ln4\pi}{2\sqrt{2\alpha}\,d}+O\!\left(\frac1d\right),
\end{equation}
where the $O(1/d)$ term is the Poisson--Dirichlet average of the gaps below the
maximum \cite[Ch.~9]{bovier2006statistical}.

On positive bags, it remains to determine the limit of $\sigma_1$.  This is
set by the growth rate of $Z_n$. Under the correspondence of
Table~\ref{si:tab:rem_dictionary}, the REM free energy
Eq.~\eqref{eq:REMfree} gives
\begin{equation}
  \lim_{d\to\infty}\frac1d\ln Z_n=\lambda\,\phi_\alpha(\lambda),
  \qquad
  \phi_\alpha(\lambda)=
  \begin{cases}
    \dfrac{\alpha}{\lambda}+\dfrac{\lambda}{2}, & \lambda<\lambda^\star\ ({\rm H}),\\[6pt]
    \sqrt{2\alpha}, & \lambda\ge\lambda^\star\ ({\rm L}).
  \end{cases}
  \label{si:eq:noise_free_energy}
\end{equation}
Comparing $\phi_\alpha(\lambda)$ to the exponent $u\rho_{\rm true}$ of the
planted instance gives
\begin{equation}
  \lim_{d\to\infty}\sigma_1=
  \begin{cases}
    1, & \rho_{\rm true}u>\phi_\alpha(\lambda)\ ({\rm R}),\\
    0, & \rho_{\rm true}u<\phi_\alpha(\lambda)\ ({\rm N}).
  \end{cases}
  \label{si:eq:retrieval_criterion}
\end{equation}

Hence $1-\sigma_1$ is exponentially small in the R
phase and $\sigma_1$ is exponentially small in the N phase. As
$d\to\infty$ we then have
\begin{align}
  \frac{1}{d^2}\mathbb E^{-}[\hat w_\lambda^2] &\to
  \begin{cases}
    \lambda^2\rho^2, & {\rm H},\\
    2\alpha\rho^2, & {\rm L},
  \end{cases}
  &
  \frac{1}{d^2}\mathbb E^{+}[\hat w_\lambda^2] &\to
  \begin{cases}
    \rho^2\rho_{\rm true}^2u^2, & {\rm R},\\
    \lambda^2\rho^2, & {\rm HN},\\
    2\alpha\rho^2, & {\rm LN}.
  \end{cases}
\end{align}

\subsubsection{Cross term}
On positive bags $w^k=\rho_{\rm true}^2d$, so
\begin{equation}\frac{1}{d^2}\mathbb E^{+}[w^k\hat w_\lambda]\approx\rho_{\rm
true}^2\,\frac1d\mathbb E^{+}[\hat w_\lambda].
\end{equation}
 On negative bags $m_n$ and
$\hat w_\lambda/d$ both have fluctuations of order $1/d$, so their
covariance is $O(1/d^2)$ and
\begin{equation}
  \frac{1}{d^2}\mathbb E^{-}[w^k\hat w_\lambda]\approx\mathbb E^{-}[m_n]\,\frac1d\mathbb E^{-}[\hat w_\lambda],
  \qquad
  \mathbb E^{-}[m_n]=\rho_{\rm true}\sqrt{2\alpha}-\rho_{\rm true}\frac{\ln(\alpha d)+\ln4\pi-2\gamma_E}{2\sqrt{2\alpha}\,d}.
\end{equation}

\subsubsection{Phase table}
Splitting the loss by bag type,
\begin{equation}
  \bar L(\lambda,u,\rho) = q\,\bar L_+(\lambda,u,\rho) + (1-q)\,\bar L_-(\lambda,u,\rho),
\end{equation}
and 
\begin{equation}
  \bar L_\pm=\frac{1}{d^2}\Big(\mathbb E^{\pm}[(w^k)^2]-2\,\mathbb E^{\pm}[w^k\hat w_\lambda]+\mathbb E^{\pm}[\hat w_\lambda^2]\Big).
\end{equation}

 Writing
$\bar w^\pm:=\lim_{d\to\infty}\hat w_\lambda/d$ on positive and negative
bags, the conditional losses therefore converge to squared residuals,
\begin{equation}
  \lim_{d\to\infty}\bar L_\pm=\Delta_\pm^2,
  \qquad
  \Delta_+=\rho_{\rm true}^2-\bar w^+,
  \qquad
  \Delta_-=\sqrt{2\alpha}\,\rho_{\rm true}-\bar w^-,
\end{equation}
where $\rho_{\rm true}^2$ and $\sqrt{2\alpha}\,\rho_{\rm true}$ are the
limits of $w^k/d$ on positive and negative bags. The output limits are
$\bar w^-=\lambda\rho$ in H and $\bar w^-=\sqrt{2\alpha}\,\rho$ in L. On
positive bags, $\bar w^+=\rho\rho_{\rm true}u$ in R, and in N it equals
$\bar w^-$. Combining the H/L criterion $\lambda\lessgtr\lambda^\star$ with
the R/N criterion of Eq.~\eqref{si:eq:retrieval_criterion} gives the four
phases of Table~\ref{tab:phases} in the main text.

\subsection{Gradient flow}
\label{si:sec:flow}

The continuous gradient flow is $\dot\lambda=-\partial_\lambda\bar L$,
$\dot\rho=-\partial_\rho\bar L$, $\dot{\bb}=-P_{\bb}\nabla_{\bb}\bar L$
with $P_{\bb}v:=v-\langle v,\bb\rangle\bb/d$ the tangent projection onto
$\sqrt d\,S^{d-1}$. Since $\bar L$ depends on $\bb$ only through $u$,
$\nabla_{\bb}\bar L=\bb_{\rm true}\partial_u\bar L$, and using
$\langle\bb_{\rm true},P_{\bb}\bb_{\rm true}\rangle=d(1-u^2)$ gives
$\dot u=-(1-u^2)\partial_u\bar L$. For $c\in\{\lambda,u,\rho\}$,
\begin{equation}
  -\partial_c\bar L = \frac2d\Big[q\,\Delta_+\,\partial_c\hat w^+ + (1-q)\,\Delta_-\,\partial_c\hat w^-\Big],
  \label{si:eq:master_gradient_rule}
\end{equation}
where $\Delta_\pm$ are the residuals in the row of
Table~\ref{tab:phases} for the current phase. Explicitly, in each phase:

\paragraph{HR (high temperature, retrieval).}
\begin{align}
  \dot\lambda &= 2(1-q)\,\rho\,\big(\sqrt{2\alpha}\,\rho_{\rm true}-\lambda\rho\big), \\
  \dot u &= 2q\,\rho\,\rho_{\rm true}^2\,(\rho_{\rm true}-\rho u)\,(1-u^2), \\
  \dot\rho &= 2q\,u\,\rho_{\rm true}^2\,(\rho_{\rm true}-\rho u) + 2(1-q)\,\lambda\,\big(\sqrt{2\alpha}\,\rho_{\rm true}-\lambda\rho\big).
\end{align}
The marked instance already dominates the softmax on positive bags, so
$\hat w^+/d$ is independent of $\lambda$ and $\dot\lambda$ is driven
purely by the negative-bag mismatch $\sqrt{2\alpha}\,\rho_{\rm true}-\lambda\rho$.

\paragraph{LR (low temperature, retrieval).}
\begin{align}
  \dot\lambda &= 0, \\
  \dot u &= 2q\,\rho\,\rho_{\rm true}^2\,(\rho_{\rm true}-\rho u)\,(1-u^2), \\
  \dot\rho &= 2q\,u\,\rho_{\rm true}^2\,(\rho_{\rm true}-\rho u) + 4\alpha(1-q)\,(\rho_{\rm true}-\rho).
\end{align}
Once the softmax on negative bags has also collapsed onto its single
largest instance, $\hat w^-/d=\sqrt{2\alpha}\,\rho$ no longer depends on
$\lambda$, so there is no gradient signal left to move the temperature:
$\dot\lambda=0$ identically. This is the freezing mechanism behind the
main text's account of how $\lambda$ evolves.

\paragraph{HN (high temperature, non-retrieval).}
\begin{align}
  \dot\lambda &= 2\rho\big[q\,\rho_{\rm true}^2+(1-q)\sqrt{2\alpha}\,\rho_{\rm true}-\lambda\rho\big], \\
  \dot u &= 0, \\
  \dot\rho &= 2\lambda\big[q\,\rho_{\rm true}^2+(1-q)\sqrt{2\alpha}\,\rho_{\rm true}-\lambda\rho\big].
\end{align}
Neither bag type has selected out a single instance, so
$\hat w^{\pm}/d=\lambda\rho$ is independent of $u$ and the overlap is
completely frozen. Since $\dot\lambda\propto\rho$ and
$\dot\rho\propto\lambda$ with the same proportionality factor,
$\lambda^2-\rho^2$ is conserved, so $(\lambda,\rho)$ moves along a
hyperbola toward the point where the bracketed residual vanishes.

\paragraph{LN (low temperature, non-retrieval).}
\begin{align}
  \dot\lambda &= 0, \\
  \dot u &= 0, \\
  \dot\rho &= 2\sqrt{2\alpha}\,\big[q\,\rho_{\rm true}^2+(1-q)\sqrt{2\alpha}\,\rho_{\rm true}-\sqrt{2\alpha}\,\rho\big].
\end{align}
Both $\lambda$ and $u$ are frozen; only $\rho$ evolves, adjusting to bring
the predicted extremal scale $\sqrt{2\alpha}\,\rho$ into agreement with
the true extremal statistics on both bag types.

\section{Robustness to noise in labeling and selection}
\label{si:sec:noise}

We address the question of whether the results are sensitive noise. Noise can be added before and after labeling, hence we consider 
\begin{equation}
  y_i=\langle\boldsymbol x_i,\bbeta_{\rm true}\rangle+\eta_i,
  \qquad
  w^k=\max_i y_i+\xi^k,
  \label{si:eq:noisy_label}
\end{equation}
where \(\eta_i\overset{\rm iid}{\sim}{\rm Normal}(0,\sigma_\eta^2d)\)
is instance-level noise and
\(\xi^k\overset{\rm iid}{\sim}{\rm Normal}(0,\sigma_\xi^2d^2)\)
is independent bag-level noise. The scalings  \(O(\sqrt d)\) are so that the noise is comparable to the background-score and
\(O(d)\) bag-label scales -- $o(\sqrt{d})$ noise will vanish in our asymptotics. 
Since neither noise effects the learned model
\(\hat w_\lambda(X^k\bbeta)\), the retrieval and H/L boundaries and the
outputs in Table~\ref{tab:phases} are unchanged and only the target
residuals can change. This reduces the calculation to finding the two
noisy limits of \(w^k/d\). 

For \(\sigma_\eta=0\), averaging over the bag level noise 
gives 
\begin{equation}
  \bar L_\pm
  =\frac{1}{d^2}\mathbb E\!\left[
    (w^{k}+\xi^k-\hat w_\lambda)^2\right]
  =\bar L_\pm^0+\sigma_\xi^2.
  \label{si:eq:xi_shift}
\end{equation}
Here \(\bar L_\pm^0\) denotes the noise-free loss for which the large-\(d\) limit is the corresponding entry of
Table~\ref{tab:phases}. The cross-term vanishes because \(\xi^k\) is
mean zero and independent of both the label and the prediction. Therefore, post-max noise
raises the population loss by a constant, leaving the gradient flow
unchanged. 

Now consider the opposite limit by setting \(\sigma_\xi=0\) and define
\begin{equation}
  \rho_{\rm true}^{(\eta)}
  :=\sqrt{\rho_{\rm true}^2+\sigma_\eta^2}.
  \label{si:eq:rho_true_eta}
\end{equation}
On a negative bag the independent signal and noise variances add, so
the scores are i.i.d.\
\({\rm Normal}(0,(\rho_{\rm true}^2+\sigma_\eta^2)d)\). Therefore
\begin{equation}
  \frac{w^{-,k}}{d}\xrightarrow{d\to\infty}
  \sqrt{2\alpha}\,\rho_{\rm true}^{(\eta)}.
\end{equation}
On a positive bag, the label of the planted instance is
\(y_1=\rho_{\rm true}^2d+\eta_1\), where
\(\eta_1\) has fluctuations of order \(\sqrt d\), and hence
\(y_1/d\to\rho_{\rm true}^2\). It competes with an \(n-1\) instance
background maximum having the same limit as above. Assuming the
identifiability condition \(\alpha<\rho_{\rm true}^2/2\),  the marked
instance remains identifiable when
\begin{equation}
  \rho_{\rm true}^2>\sqrt{2\alpha}\,\rho_{\rm true}^{(\eta)}
  \quad\Longleftrightarrow\quad
  \sigma_\eta<\sigma_\eta^{\rm crit}
  :=\rho_{\rm true}\sqrt{\frac{\rho_{\rm true}^2}{2\alpha}-1}.
  \label{si:eq:sigma_eta_crit}
\end{equation}
Above this threshold, positive and negative labels have the same
leading-order limit, so the labels no longer identify the planted
instance. We restrict the following statements to the identifiable
regime below the threshold. There the positive-bag target remains
\(\bar y^+=\rho_{\rm true}^2\), while
\(\bar y^-=\sqrt{2\alpha}\rho_{\rm true}^{(\eta)}\). Hence only the negative-bag residual \(\Delta_-\) in
Table~\ref{tab:phases} changes:
\begin{equation}
  \Delta_-\longrightarrow
  \begin{cases}
    \sqrt{2\alpha}\rho_{\rm true}^{(\eta)}-\lambda\rho & \text{(HR, HN)},\\
    \sqrt{2\alpha}\big(\rho_{\rm true}^{(\eta)}-\rho\big) & \text{(LR, LN)}.
  \end{cases}
  \label{si:eq:residuals_eta}
\end{equation}
Replacing \(\Delta_-\) by these expressions in
Sec.~\ref{si:sec:flow} gives the complete noisy flow. In particular,
the phase boundaries, \(\dot u=0\) in HN/LN, and \(\dot\lambda=0\) in
LR/LN are unchanged: these properties follow from the pooled
prediction, whose dependence on \((\lambda,u,\rho)\) is unaffected by
the noisy targets.

Conditioning on the pre-max noise does not effect the argument leading to
Eq.~\eqref{si:eq:xi_shift}. Therefore, the effects simply add and instance level noise
noise makes the substitutions in Eq.~\eqref{si:eq:residuals_eta}, while
independent mean-zero post-max noise contributes only the constant
\(\sigma_\xi^2\) to the loss.

\section{Connection to pattern retrieval in associative memory}
\label{si:sec:hopfield}

The connection to modern Hopfield networks is the common softmax
competition between one distinguished pattern and exponentially many
background patterns. We write the associative-memory problem in the
notation of the MIL model. The \(n\) instances
\(\boldsymbol x_1,\dots,\boldsymbol x_n\in\mathbb R^d\) of a bag \(X\)
play the role of stored patterns, the query \(\bbeta\) plays the role of
the network state, and \(\lambda\) is the inverse temperature. The
continuous Hopfield energy is
\begin{equation}
  E(\bbeta)
  =-\frac{1}{\lambda}\log\sum_{i=1}^n
    e^{\lambda\langle\boldsymbol x_i,\bbeta\rangle}
    +\frac12\|\bbeta\|_2^2.
  \label{si:eq:hopfield_energy}
\end{equation}
A unit gradient-descent step gives the vector-valued attention readout
\begin{equation}
  \bbeta_{t+1}=\boldsymbol m_\lambda(\bbeta_t;X)
  :=\sum_{i=1}^n a_i(\bbeta_t)\,\boldsymbol x_i,
  \qquad
  a_i(\bbeta)
  =\frac{e^{\lambda\langle\boldsymbol x_i,\bbeta\rangle}}
  {\sum_j e^{\lambda\langle\boldsymbol x_j,\bbeta\rangle}}.
  \label{si:eq:hopfield_update}
\end{equation}
Associative-memory retrieval means that iterating this map from a state
near \(\boldsymbol x_1\) converges to a fixed point exponentially close
to that stored pattern
\cite{ramsauer2020hopfield,lucibello2024exponential}.

The MIL predictor uses these attention weights but has a scalar output.
The definition in main-text Eq.~\eqref{eq:softmax_predictor} gives the identity
\begin{equation}
  \hat w_\lambda(X\bbeta)
  =\langle\bbeta,\boldsymbol m_\lambda(\bbeta;X)\rangle.
  \label{si:eq:mil_hopfield_projection}
\end{equation}
Our pooling operation is therefore a projection of the Hopfield readout
onto the query. In a positive bag, the weight
on the planted instance is set by a signal-versus-noise comparison that
mirrors associative-memory retrieval. Its exponential weight competes
with the REM partition function \(Z_n\) in Eq.~\eqref{si:eq:sigma_truth}.

\section{Extension to a classification loss}
\label{si:sec:classification}

Here we show how the results generalize to a classification problem. 
Every quantity in Sec.~\ref{si:sec:asymptotics} up through
Eq.~\eqref{si:eq:retrieval_criterion} and Table~\ref{tab:phases}'s
$\bar w^{\pm}$ column is a statement about the large-$d$ limit of the
pooled score $\hat w_\lambda(X\bbeta)/d$ alone -- in particular, these results never reference the loss function. For
\emph{any} loss $\ell(z,\hat w_\lambda)$ built from the normalized pooled
score $\hat w_\lambda/d$ and a label $z$, the chain rule gives
\begin{equation}
  -\partial_c\bar L = q\,R^+(\bar w^+)\,\partial_c\bar w^+ + (1-q)\,R^-(\bar w^-)\,\partial_c\bar w^-,
  \qquad c\in\{\lambda,u,\rho\},
  \label{si:eq:general-master-rule}
\end{equation}
where $R^{\pm}:=-\partial_{\hat w}\ell(z^{\pm},\hat w)|_{\hat w=\bar w^{\pm}}$
is a scaler depending only on the loss and the
label, and $\partial_c\bar w^{\pm}$ is the same architecture-determined
Jacobian entry read off Table~\ref{tab:phases} in either case. The
phase boundaries and parameter dependence of the pooled scores follow
entirely from the three asymptotic forms $\rho\rho_{\rm true}u$,
$\lambda\rho$, and
$\sqrt{2\alpha}\rho$. Thus, any loss built from the same pooled score
inherits the same four phases: $u$ is frozen in the non-retrieval phases
HN and LN, while $\lambda$ is frozen in the low-temperature phases LR and
LN because the pooled scores are independent of $\lambda$. The choice of
loss affects only the residuals driving the flow, and hence the fixed
points within each phase.

\subsection{The logistic (cross-entropy) case}

Concretely, take the label $z^k=\mathbf 1\{\bbeta_{\rm true}\in X^k\}$
(the same containment event that already defines positive/negative bags),
model $\hat p_\lambda=g(\hat w_\lambda/d-\tau)$ with $g$ the logistic
sigmoid and $\tau$ a fixed threshold, and minimize the cross-entropy
$\ell(z,p)=-z\ln p-(1-z)\ln(1-p)$. The standard fact
$\partial_s\ell(z,g(s))=g(s)-z$ gives
\begin{equation}
  R^+ = 1-\bar p^+ \in(0,1), \qquad R^- = -\bar p^- \in(-1,0), \qquad \bar p^{\pm}:=g(\bar w^{\pm}-\tau),
\end{equation}
in place of the regression problem's linear residuals $2\Delta_\pm$.
We now substitute these residuals into
Eq.~\eqref{si:eq:general-master-rule} phase by phase.

\paragraph{HN and LN.}
The fixed point $\bar p=q$ remains closed form:
$\lambda\rho=\tau+\ln\frac{q}{1-q}$ in HN and
$\rho^\star=(\tau+\ln\frac{q}{1-q})/\sqrt{2\alpha}$ in LN, provided
$\tau+\ln\frac{q}{1-q}>0$. Otherwise, the flow runs to the
$\lambda\rho=0$ boundary. The conserved hyperbola $\lambda^2-\rho^2$ of
the HN phase is unchanged. 

\paragraph{LR.}
The residual $R^+=1-\bar p^+$ is strictly positive, so $\dot u>0$
throughout the phase and $u$ runs to the boundary $u=1$ rather than to
an interior overlap. The stationarity condition for $\rho$,
$q(1-\bar p^+)\,\rho_{\rm true}u=(1-q)\sqrt{2\alpha}\,\bar p^-$, is
transcendental because the residuals are now sigmoidal in the score.

\paragraph{HR.}
In HR, the cross-entropy gradient gives
\begin{equation}
  \dot\lambda=-(1-q)\rho\,\bar p^-<0,
\end{equation}
so $\lambda$ decreases. However, the trajectory cannot remain in the
retrieval phase as $\lambda\to0$, because $\phi_\alpha(\lambda)$ diverges
while retrieval requires
$\rho_{\rm true}u>\phi_\alpha(\lambda)$. The trajectory therefore crosses
the retrieval boundary at
\begin{equation}
  \lambda_c(u)=\rho_{\rm true}u-
  \sqrt{(\rho_{\rm true}u)^2-2\alpha}
\end{equation}
and enters HN, provided $\rho$ remains positive. The absence of a finite optimum is familiar from
cross-entropy on separable data \cite{soudry2018implicit}. The
consequence for the dynamics is specific, however: the route through HR
into LR by which the regression flow reaches $\lambda^\star$ is closed.

Under cross-entropy, the four phases and the freezing of $u$ outside
retrieval and of $\lambda$ in the low-temperature phases remain
unchanged. However, the tuning of $\lambda$ to $\lambda^\star$ is lost.
In the regression problem, the target contains the extremal scale
$\sqrt{2\alpha}\,\rho_{\rm true}$, which drives $\lambda$ upward through
HR. A binary label contains no corresponding scale, so $\lambda$ instead
decreases in HR and the flow approaches fixed points determined by the
classification threshold $\tau$, rather than by $\lambda^\star$.

\section{Numerical tests of the gradient-flow limit}
\label{si:sec:numerics}

The figures below compare the finite-size empirical gradient against the
asymptotic field of Eq.~\eqref{si:eq:master_gradient_rule}, with the
residuals of Table~\ref{tab:phases}. The gradient is
evaluated at a point in $(\lambda,u,\rho)$, so this doesn't incorporate any accumulated bias of the learned parameters, which can in principle lead to higher dimensional dynamics in the $d \to \infty$ limit.

\begin{figure}[h]
  \centering
  {\raggedright (a)\par}
  \includegraphics[width=\textwidth,trim=0 0 0 23bp,clip]{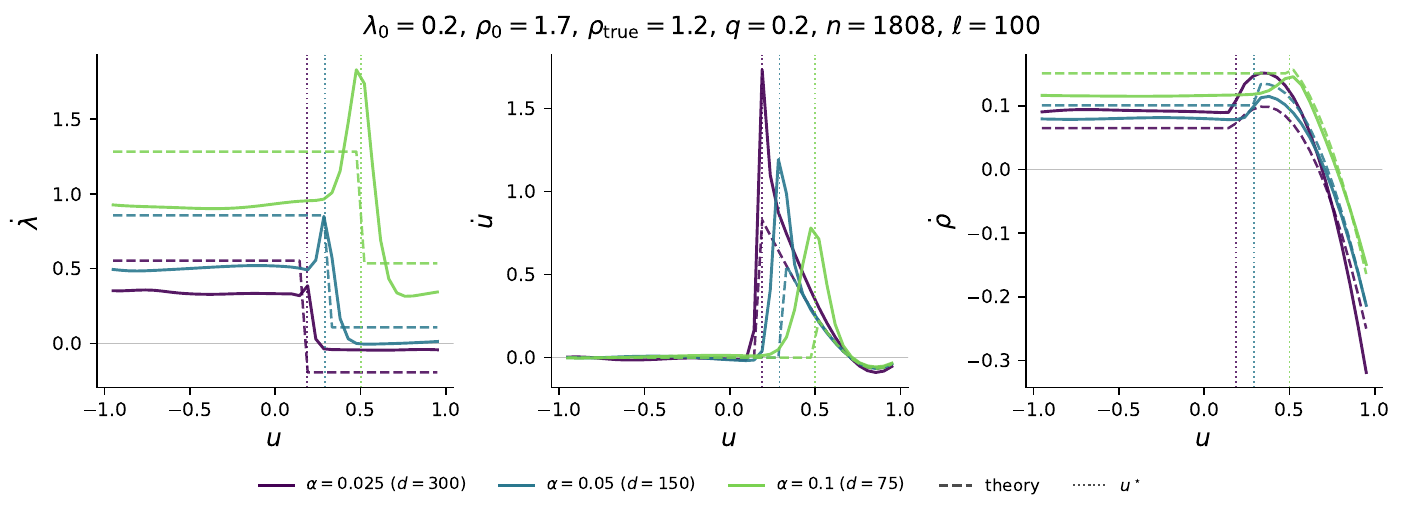}\\[4pt]
  {\raggedright (b)\par}
  \includegraphics[width=\textwidth,trim=0 0 0 23bp,clip]{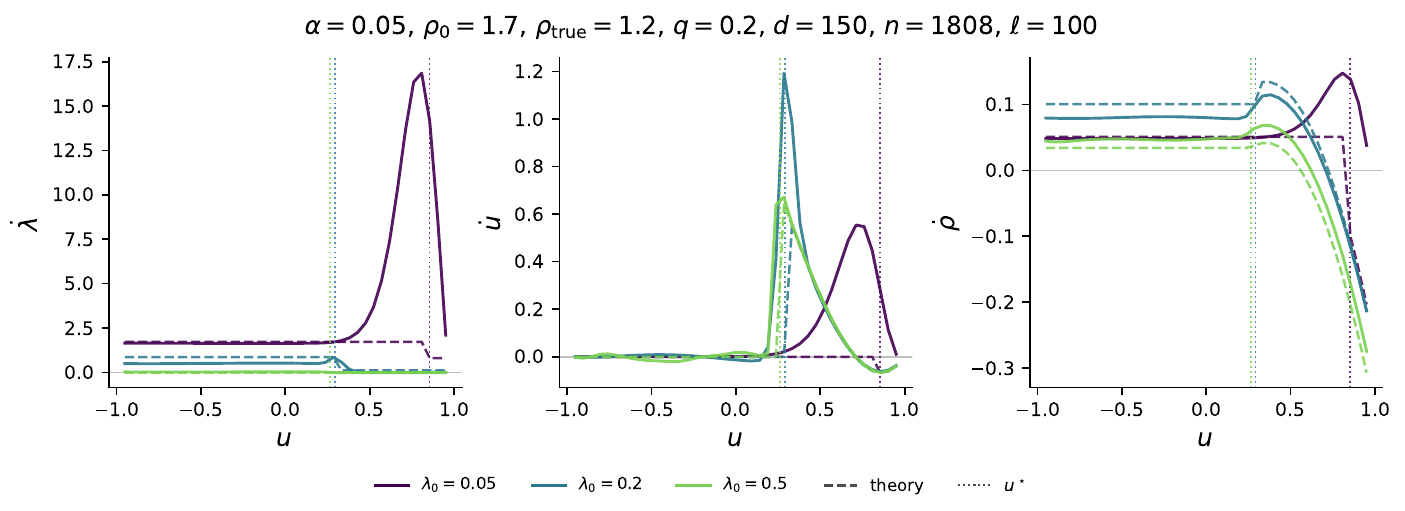}
  \caption{Empirical $(\dot\lambda,\dot u,\dot\rho)$ (solid) and the
    asymptotic prediction (dashed, same color) against the overlap $u$,
    at $\rho_0=1.7$, $\rho_{\rm true}=1.2$, $q=0.2$, $\ell=100$ and
    $n=1808$ throughout. (a) Three values of $\alpha$ at
    $\lambda_0=0.2$, with $d=300$, $150$ and $75$. (b) Three initial
    temperatures at $\alpha=0.05$ and $d=150$. Dotted vertical lines mark
    $u^\star$.}
  \label{si:fig:alpha_lam0}
\end{figure}

Figure~\ref{si:fig:alpha_lam0} sweeps the overlap $u$ at $\rho_0=1.7$,
$\rho_{\rm true}=1.2$, $q=0.2$ and $\ell=100$. Row (a) varies $\alpha$ at
$\lambda_0=0.2$, with $d=300$, $150$ and $75$ so that $n=1808$ throughout.
Row (b) fixes $\alpha=0.05$ and varies the initial temperature.
$\lambda^\star=0.316$ in row (b), so $\lambda_0=0.05$ and $0.2$ lie in the
high-temperature phase and $\lambda_0=0.5$ in the low-temperature phase,
where $\bar w^\pm$ have no $\lambda$ dependence and the asymptotic
$\dot\lambda$ is identically zero. The retrieval threshold is
$u^\star=0.854$, $0.292$ and $0.264$ respectively.

The other sweep is over $\lambda_0$.  Over $\lambda_0\in[0.1,0.4]$ a fixed $u$ stays in R throughout
only if $u>0.458$, and in N throughout only if $u<0.264$. The two panels
take $u=0.7$ and $u=0.1$.

\begin{figure}[h]
  \centering
  \includegraphics[width=0.9\textwidth,trim=0 0 0 22bp,clip]{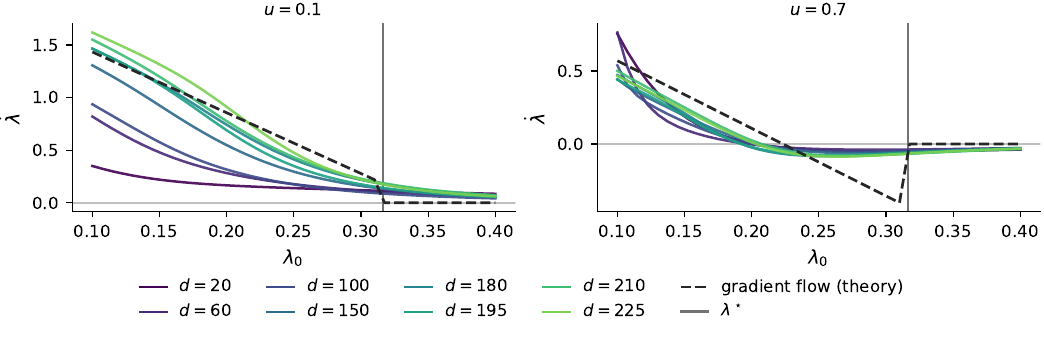}
  \caption{$\dot\lambda$ against $\lambda_0$ at fixed overlap, for
    increasing $d$ (with $n=e^{\alpha d}$), and the asymptotic prediction
    (dashed). Parameters are $\alpha=0.05$, $\rho_0=1.7$,
    $\rho_{\rm true}=1.2$, $q=0.2$, $\ell=100$. Left: $u=0.1$, in N throughout. Right: $u=0.7$, in R
    throughout. The vertical line is $\lambda^\star=0.316$.}
  \label{si:fig:lam_sweep}
\end{figure}

\begin{figure}[h]
  \centering
  \includegraphics[width=\textwidth]{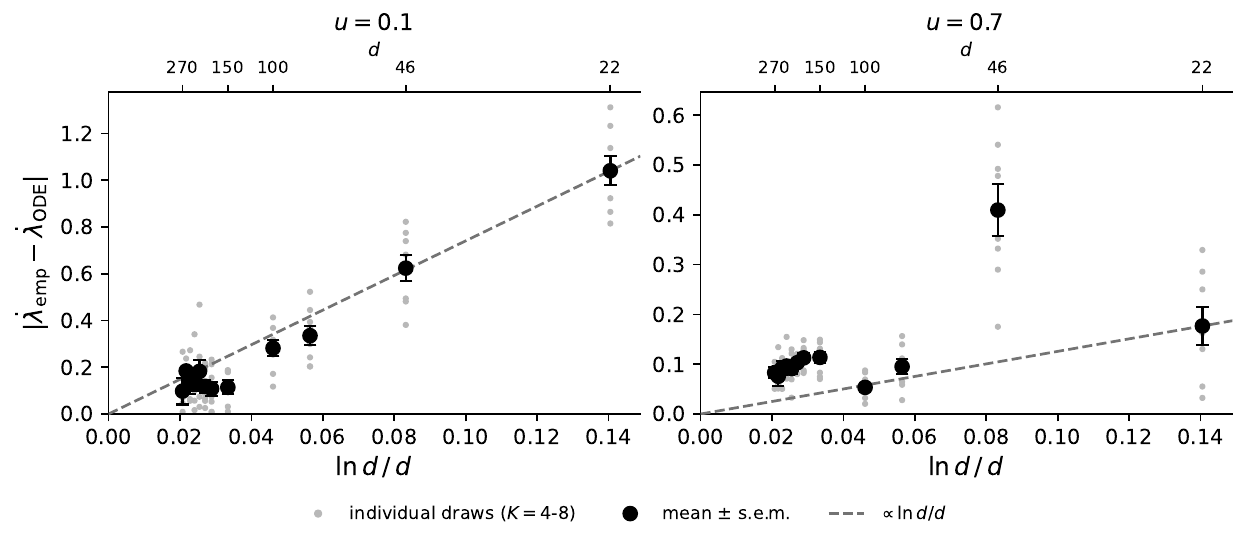}
  \caption{Error in the initial $\dot\lambda$ vs.\ $\ln d/d$, for
    $u=0.1$ and $u=0.7$, at $\lambda_0=0.1$, $\rho_0=1.7$, $\alpha=0.05$,
    $\rho_{\rm true}=1.2$, $q=0.2$, $\ell=100$. $d$ runs from 22 to 270
    ($n=3$ to $7.3\times10^5$), marked on the top axis. Grey points are
    individual datasets, black their mean with the standard error: eight
    datasets per $d$ up to $d=225$, four above. Dashed line: $\ln d/d$
    through the origin and the smallest-$d$ point.}
  \label{si:fig:gradient_convergence}
\end{figure}

\section{LLM disclosure}
\label{si:sec:llm}

Anthropic's Claude models (Claude Opus 5, Opus 5.5, Sonnet 5 and Fable 5),
accessed through the Claude Code agent, were used for coding, for
mathematical derivations, and assisting in writing the text. Substantial edits
to the text (anything beyond formatting and grammar) and the generation of
large sections of code were managed with git. In particular, we posted a issues
on GitHub describing the required change and prompted Claude Code to address
them in pull requests which we reviewed. This
workflow allowed us to track and review all substantial LLM-generated
material.

\end{document}